\documentclass[letterpaper, 10 pt, conference]{ieeeconf} 
\IEEEoverridecommandlockouts
\usepackage{graphicx} 
\let\labelindent\relax
\usepackage{enumitem}
\usepackage{hyperref}
\usepackage{amsmath}
\usepackage{amssymb}
\usepackage{algorithm}
\usepackage{algorithmic}
\usepackage{color}
\usepackage{authblk}
\usepackage{float}

\title{\textbf{Active Real-World Factor-Based Evaluation \\for Generalist Robot Policies}}
\author[1]{Andrew Liao}
\author[1]{Hanchen Cui}
\author[1]{Karthik Desingh}
\author[1]{Aryan Deshwal}
\affil[1]{University of Minnesota Twin Cities}

\begin{document}

\maketitle

\begin{abstract} 
Generalist robot manipulation policies trained on large, diverse datasets have shown remarkable promise across a wide range 
of tasks. However, rigorously evaluating these policies remains a fundamental challenge. Real-world performance depends on a 
large combinatorial space of task factors including object poses and camera viewpoints, making full, exhaustive evaluation 
intractable. Additionally, real hardware evaluation is slow and resource-intensive, so current practice is to use narrow test suites that can miss critical failure modes and misrepresent true deployment readiness. We propose 
an active evaluation framework that addresses this challenge by treating policy evaluation as a sequential 
experimental design problem. Our approach fits a probabilistic surrogate model over a structured space of task factors and 
adaptively selects evaluation configurations to maximize information gain over the policy's performance distribution, allowing 
for sample-efficient characterization of policy behavior across unseen conditions and a systematic identification of 
failure-prone regions. We conduct 2331 real-world evaluations across 3 tasks with 
3 factor variations and find that our approach typically saves the evaluator at least 20-40\% of trials compared to typical random testing.
\end{abstract}

\section{Introduction}
Recent advances in foundation models across vision, language, and multimodal domains~\cite{radford2021learningtransferablevisualmodels, brown2020languagemodelsfewshotlearners, liu2023visualinstructiontuning} have inspired a parallel surge in generalist robot manipulation policies. These models use similar architectures and training paradigms to perform various manipulation and navigation tasks~\cite{brohan2023rt2visionlanguageactionmodelstransfer, kim2024openvlaopensourcevisionlanguageactionmodel, black2026pi0visionlanguageactionflowmodel}, demonstrating remarkable breadth and tackling long-horizon and dexterous tasks that were previously out of reach for robotic systems.

\begin{figure}
    \centering
    \includegraphics[width=1\linewidth]{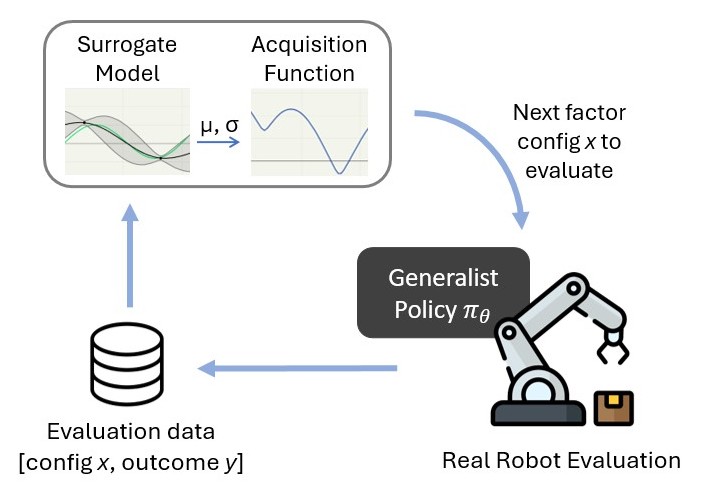}
    \caption[Active evaluation framework]{Illustration of our proposed approach which  uses Bayesian active learning to adaptively select the most informative task configurations for real-world robot evaluation. This enables sample-efficient estimation of the generalist policy's performance.}
    \label{fig:active_testing}
    \vspace{-2ex}
\end{figure}

Despite this progress, generalist robot policies continue to struggle with generalization in ordinary, real-world environments. Real-world deployment scenarios are characterized by a large combinatorial space of task factors, including camera viewpoints, language instruction phrasings, initial robot and object states, table heights, and more. Policies often fail or exhibit shortcut behaviors when evaluated outside their training distribution~\cite{fei2025liberoplusindepthrobustnessanalysis, xing2025shortcutlearninggeneralistrobot}. Comprehensively characterizing a policy's performance therefore requires evaluating it across the full range of these factors and their possible values. However, this is computationally intractable, and physical robot experiments further compound the problem: real-hardware manual evaluation is slow, resource-intensive, and susceptible to many evaluator biases~\cite{kressgazit2024robotlearningempiricalscience}. As a result, typical evaluations in the literature are simple and often narrow in scope, using binary success rate as opposed to continuous (oversimplified), usually performing $<$15 evaluations per task (no statistical rigor), and do not faithfully represent the breadth of conditions an end user would encounter (only a few variations in the domain, or environment, factors). This can mask critical failure modes and inflate reported performance, obscuring the true deployment readiness of a policy. Although high-fidelity simulators have partially bridged these gaps~\cite{makoviychuk2021isaacgymhighperformance, Todorov2012MuJoCoAP}, allowing for more evaluations and domain variations, visual and physics discrepancies remain, making real-world evaluation the gold standard. Additionally, since compute is finite, even simulation may not suffice at times.

In this work, we address this gap by proposing a sample-efficient, factor-based active evaluation framework for generalist robot manipulation policies on real hardware. Our evaluation approach models policy performance over a structured space of task factors using a probabilistic surrogate model, and actively selects the next factor configurations to maximize information gain over the performance distribution across the design space. This enables us to efficiently and accurately characterize policy behavior across unseen conditions and identify failure-prone regions. With over 700 ground truth real-world evaluations per task across three tasks, we demonstrate the benefits of active testing compared to standard random testing. With an evaluation budget of 100 trials, our active framework typically saves the evaluator 20-40 trials (20-40\% less work). We also show that using surrogate models to model policy performance across the factor design space leads to good predictive accuracy, allowing us to perform---in fewer trials---typical performance analyses like characterizing generalization of the policy to different factors.

\section{Related Work}
\subsection{Generalist Robot Policies}
The primary aspiration in the field of robot learning is to build a solution to performing arbitrarily difficult tasks in robot manipulation or locomotion. "Generalist robot policies" attempt to do this with one unified model, usually taking in visual observations and natural language instructions to produce robot actions~\cite{ma2026surveyvisionlanguageactionmodelsembodied}. Early work before the advent of large foundation models essentially trained large-scale behavior cloning models \cite{jang2022bczzeroshottaskgeneralization, reed2022generalistagent, brohan2023rt1roboticstransformerrealworld}. More recent work have leveraged large open-source datasets and pretrained Vision-Language models (VLMs) as backbones to construct Vision-Language-Action models (VLAs)~\cite{ embodimentcollaboration2025openxembodimentroboticlearning, khazatsky2025droidlargescaleinthewildrobot, brohan2023rt2visionlanguageactionmodelstransfer, kim2024openvlaopensourcevisionlanguageactionmodel, black2026pi0visionlanguageactionflowmodel}, achieving promising advances in visual generalization, multi-task proficiency, and dexterity.

\subsection{Benchmarks for Generalist Robot Policies}
Benchmarks evaluating generalist robot policies focus on their generalization to out-of-distribution conditions (as opposed to other measures like repeatability, speed, or safety). Several works have defined this as robustness to various human-interpretable perturbation factors (e.g., lighting, object placement, table texture, language instruction variations) \cite{gao2026taxonomyevaluatinggeneralistrobot,parekh2024investigatingroleinstructionvariety} and multiple benchmarks have been developed to measure this
\cite{liu2023liberobenchmarkingknowledgetransfer, mees2022calvinbenchmarklanguageconditionedpolicy, fei2025liberoplusindepthrobustnessanalysis, pumacay2024colosseumbenchmarkevaluatinggeneralization, xie2023decomposinggeneralizationgapimitation, zhang2024vlabenchlargescalebenchmarklanguageconditioned}. These studies have assisted in pinpointing weaknesses and strengths of generalist robot policies; for instance, these policies are usually more sensitive to changes in camera viewpoint and initial robot and object states than to changes in lighting and background texture. Our work focuses on how to do factor-based evaluation efficiently.

\subsection{Sample-efficient Robot Policy Evaluation}
A practical challenge in robot policy evaluation---especially on real hardware---is that each trial can be time-expensive and therefore evaluators need to work with a tight budget.  Although research in building better simulations \cite{Koenig2004DesignAU, Todorov2012MuJoCoAP, makoviychuk2021isaacgymhighperformance}, which are much more scalable, have somewhat bridged the gap, visual and physics gaps still remain and real evaluation remains the gold standard. Some recent works have investigated real-to-sim evaluation, showing promising correlation between results in simulation and real \cite{li2024evaluatingrealworldrobotmanipulation, torne2024reconcilingrealitysimulationrealtosimtoreal,zhang2025realtosimrobotpolicyevaluation, sedlacek2025realmrealtosimvalidatedbenchmark, jain2025polarisscalablerealtosimevaluations, abouchakra2025realissimbridgingsimtorealgap, yu2025real2render2realscalingrobotdata, jangir2026robotarenainftyscalablerobot}; however, these works are not yet visually indistinguishable from real. In any case, whether simulation or real evaluation prevails, potential time and compute savings necessitate efficient evaluation methods.

Sample-efficient evaluation of robot policies has been studied through several methods. These include effort-aware perturbations to the factor configuration to limit switching costs \cite{anwar2024contrastsetsevaluatinglanguageguided}, combining real and simulation evaluations \cite{badithela2025reliablescalablerobotpolicy}, and early stopping with risk limits \cite{snyder2025imitationlearningpolicybetter}. However, these works are not relevant to the setting we tackle in this work: single policy, real-world performance estimation---as opposed to performance \textit{comparison}---on a fixed evaluation set of configurations.

From the broader machine learning literature, active testing, the evaluation analogue of Bayesian active learning \cite{chaloner1995bayesian,seo2000gaussian}, attempts to reduce evaluation cost by selecting the next best input from test data to label \cite{Sawade2010ActiveRE, kossen2021activetestingsampleefficientmodel, kossen2022activesurrogateestimatorsactive, li2020activetestingunbiasedevaluation}. In robotics, cost-aware active testing has been explored for evaluating robot policy performance for different policy-task combinations in simulation \cite{anwar2025efficientevaluationmultitaskrobot}. This work is the most relevant to our setting; in comparison, we similarly use active testing for evaluating policy performance but instead of using the space of policy-task combinations in simulation, we use the space of environmental factors (camera viewpoint, table height, object position) for real-robot evaluation under strict budgets (100 trials), and compare a much larger assortment of surrogate models and acquisition functions (6 combinations versus 1).

\section{Methodology}
In this section, we describe our approach for sample-efficient factor-based evaluation of generalist robot policies in the real world. The key challenge is that evaluating a robot policy, under a given task configuration, requires a physical experiment on real hardware which makes exhaustively evaluating the policy across the design space prohibitively expensive. Our key idea is to leverage Bayesian active testing where we build a probabilistic surrogate model of the policy's performance and use an acquisition function to intelligently select which factor configurations to evaluate next based on the surrogate's predictive mean and uncertainty. This approach is illustrated by Figure \ref{fig:active_testing}. We first describe the evaluation factor design space and our performance estimation objective, and then describe the full active testing procedure. 

\subsection{Problem Formulation}
In this section, we describe the problem formulation which includes the design space of task configurations over which the robot policy will be evaluated, along with the evaluation function that we want to learn as well as possible.

{\bf \noindent Design Space:} Let $\pi_\theta$ denote a generalist robot policy that maps visual observations and language instructions to robot actions. We consider the problem of evaluating $\pi_\theta$ across a design space $\mathcal{X}$ of real-world task configurations. Each configuration is defined by a set of $d$ task factors $\mathcal{F}_1, \mathcal{F}_2, \ldots, \mathcal{F}_d$, where each factor $\mathcal{F}_i$ specifies a dimension of variation relevant to deployment (e.g., object position on the table, table height, camera viewpoint). An evaluation point $\mathbf{x} \in \mathcal{X}$ is a specific assignment of values to all factors:
$$\mathbf{x} = (x_1, x_2, \ldots, x_d) \in \mathcal{F}_1 \times \mathcal{F}_2 \times \cdots \times \mathcal{F}_d$$
where each $x_i \in \mathcal{F}_i$ can be discrete or continuous depending on the factor. In the context of the manipulation tasks considered in this work, we study three factors: object position on a $11 \times 11$ discretized grid of the table surface, table heights of $\{1, 2, 3\}$ inches above the robot platform, and scene camera viewpoint from $\{\text{back}, \text{right}, \text{back-right}\}$. This creates a design space containing hundreds of valid configurations per task. 

{\bf \noindent Policy Performance Evaluation and Objective:} We assume an unknown performance function $f: \mathcal{X} \to \mathbb{R}$ that maps a task configuration to a scalar score, or outcome, reflecting the policy's performance on a task-specific continuous scale (e.g., ranging from 0 for complete failure to 10 for full success). Each evaluation of $f$ is expensive since it requires a physical execution of the policy $\pi_\theta$ on real hardware under configuration $\mathbf{x}$. 

Our goal is to estimate the full real-world performance distribution $\{f(\mathbf{x})\}_{\mathbf{x} \in \mathcal{X}}$ across the entire factor design space with minimal real-hardware robot performance evaluations.

\subsection{Active Testing}

Bayesian active testing provide a principled framework to achieve our goal of reducing uncertainty about the unknown performance function $f$ over the design space as efficiently as possible. In other words, we conduct evaluations that are maximally informative about robot policy across the design space. The key idea is to learn a probabilistic surrogate model that captures a distribution over $f$ by explicitly capturing epistemic uncertainty and using it to guide selection of factor configurations with an acquisition function that scores the expected reduction in uncertainty of candidates.

\subsubsection{Surrogate Models}
A key requirement for an effective Surrogate Model (SM) in our problem is that it must provide well-calibrated uncertainty estimates while learning from a small number of real-world evaluations. We investigate several families of probabilistic models in this work that satisfy these requirements. 

{\em \noindent SM-I).} Gaussian Processes (GPs) are an effective model that provides principled Bayesian uncertainty quantification. A GP defines a distribution over functions and is fully specified by a mean function $\mu_0: \mathcal{X} \to \mathbb{R}$ and a kernel (covariance) function $k: \mathcal{X} \times \mathcal{X} \to \mathbb{R}$. We write $f \sim \mathcal{GP}(\mu_0, k)$. Given a dataset $\mathcal{D}_t = \{(\mathbf{x}_i, y_i)\}_{i=1}^t$ of evaluated configurations with noisy observations $y_i = f(\mathbf{x}_i) + \epsilon_i$ where $\epsilon_i \sim \mathcal{N}(0, \sigma_n^2)$, the GP posterior at any candidate point $\mathbf{x}$ is obtained in closed form:
\begin{align}
    \mu_t(\mathbf{x}) &= \mathbf{k}_t(\mathbf{x})^\top (\mathbf{K}_t + \sigma_n^2 \mathbf{I})^{-1} \mathbf{y}_t\\
    \sigma_t^2(\mathbf{x}) &= k(\mathbf{x}, \mathbf{x}) - \mathbf{k}_t(\mathbf{x})^\top (\mathbf{K}_t + \sigma_n^2 \mathbf{I})^{-1} \mathbf{k}_t(\mathbf{x})
\end{align}

where $\mathbf{y}_t = (y_1, \ldots, y_t)^\top$, $\mathbf{K}_t \in \mathbb{R}^{t \times t}$ is the Gram matrix with entries $[\mathbf{K}_t]_{ij} = k(\mathbf{x}_i, \mathbf{x}_j)$, and $\mathbf{k}_t(\mathbf{x}) = (k(\mathbf{x}_1, \mathbf{x}), \ldots, k(\mathbf{x}_t, \mathbf{x}))^\top$.  The posterior mean $\mu_t(\mathbf{x})$ provides a prediction of $f(\mathbf{x})$, while the posterior variance $\sigma_t^2(\mathbf{x})$ quantifies epistemic uncertainty. 

The kernel function \cite{smola1998learning} is the key component of a GP model; it determines which inputs are considered similar and thus expected to have correlated outputs. 
In essence, it encodes prior assumptions about the smoothness and structure of $f$. 
Some popular choices are the Matern-5/2 and Radial Basis Function (RBF) kernels.
We employ a Radial Basis Function (RBF) kernel with Automatic Relevance Determination (ARD) over the full design space:
\begin{align}
    k(\mathbf{x}, \mathbf{x}') &=  \sigma^2 \exp\left( -r^2  \right), \\ \quad r &= \sqrt{\sum_{i=1}^{d} \frac{(x_i - x_i')^2}{\ell_i^2}}
\end{align}
where $\ell_i > 0$ is a per-factor lengthscale and $\sigma^2$ is the output variance.
The ARD parameterization specifies a separate lengthscale $\ell_i$ for each factor dimension, allowing the model to automatically learn which factors most strongly influence performance. All kernel hyperparameters $\{\sigma^2, \ell_1, \ldots, \ell_d\}$ are learned by maximizing the marginal log-likelihood of the GP on the evaluation data accumulated in each round.

\vspace{1ex}

{\em \noindent SM-II).} Mixture Density Networks (MDNs) are neural networks with a mixture of Gaussians as the output layer to model $p(y \mid \mathbf{x})$ directly. The network takes a configuration $\mathbf{x}$ as input and outputs the parameters of a Gaussian mixture including mixing weights $\{\pi_m\}$, means $\{\mu_m\}$, and variances $\{\sigma_m^2\}$ for $m = 1, \ldots, M$ components:
\begin{align}
    p(y \mid \mathbf{x}) = \sum_{m=1}^{M} \pi_m(\mathbf{x}) \, \mathcal{N}\bigl(y \mid \mu_m(\mathbf{x}), \sigma_m^2(\mathbf{x})\bigr)
\end{align}

In accordance with recent work, we use Monte-Carlo Dropout to estimate uncertainty for the MDN \cite{anwar2025efficientevaluationmultitaskrobot, gal2016dropoutbayesianapproximationrepresenting}. 
Monte-Carlo Dropout performs multiple forward passes through the model, calculates the final mean prediction by averaging the mean predictions of these passes, and calculates uncertainty by taking the variance of these mean predictions.

The predictive mean and variance are then:
\begin{align}
    \hat{\mu}(\mathbf{x}) &= \frac{1}{S} \sum_{s=1}^{S} \mu^{(s)}(\mathbf{x}) \\ \hat{\sigma}^2(\mathbf{x}) &= \frac{1}{S} \sum_{s=1}^{S} \bigl[\sigma^{2(s)}(\mathbf{x}) + \mu^{(s)}(\mathbf{x})^2\bigr] - \hat{\mu}(\mathbf{x})^2
\end{align}
where $\mu^{(s)}$ and $\sigma^{2(s)}$ are the mixture mean and variance from the $s$-th stochastic forward pass, and $S$ is the number of passes.

\vspace{1ex}

{\em \noindent SM-III).} Deep Ensembles provide uncertainty estimates by training $E$ independent neural networks with different random initializations on the same dataset and aggregating their predictions. Each ensemble member $e$ produces a predictive distribution $\mathcal{N}(y \mid \mu_e(\mathbf{x}), \sigma_e^2(\mathbf{x}))$, and the ensemble-level prediction is a mixture of Gaussians:
\begin{align}
    p(y \mid \mathbf{x}) = \frac{1}{E} \sum_{e=1}^{E} \mathcal{N}\bigl(y \mid \mu_e(\mathbf{x}), \sigma_e^2(\mathbf{x})\bigr)
\end{align}

It should be noted that our surrogate model differs from those used in active testing works in other domains in one important respect: 
it is not trained on the data used to train the original policy. In typical active testing settings for classifiers, the surrogate can 
leverage the target model's training data to bootstrap predictions. This does not apply in our robotics setting because the policy 
$\pi_\theta$ predicts actions while the surrogate model predicts performance outcomes---fundamentally different quantities. It does not make sense to use the training data of task demonstrations (typically successes only) to bootstrap the surrogate model that predicts performance outcome. Furthermore, because it is not common practice to release training data alongside pretrained robot policies, there may not even be training data available. Therefore, the surrogate is trained exclusively on the real-world evaluation data, making data efficiency all the more critical.

\vspace{2ex}

\subsubsection{Acquisition Functions}
Given the surrogate model, we select the next evaluation configuration by optimizing an Acquisition Function (AF) which captures how informative a candidate would be about the unknown performance values at all remaining configurations. We describe the several popular acquisition functions from Bayesian active learning we employ in this work:

{\em \noindent AF-I). Posterior Standard Deviation (PSD):} The simplest exploration-driven design criterion selects the configuration where the surrogate model is most uncertain i.e.,  point of maximum posterior uncertainty $\max_x \sigma_t(\mathbf{x})$. 

{\noindent \em AF-II). Negative Integrated Posterior Variance (NIPV):} 
While PSD scores candidates based on local properties of the posterior at a single point, the Negative Integrated Posterior Variance acquisition function \cite{861310} takes a global perspective and selects the candidate that minimizes the expected posterior variance integrated over a reference distribution $\hat{p}(\mathbf{x})$ on the design space:
\begin{align}
    \alpha_{\text{NIPV}}(\mathbf{x}) = -\int_{\mathcal{X}} \sigma_{t+1}^2(\mathbf{x}' \mid \mathbf{x}) \, \hat{p}(\mathbf{x}') \, d\mathbf{x}'
\end{align}
where $\sigma_{t+1}^2(\mathbf{x}' \mid \mathbf{x})$ is the posterior variance at $\mathbf{x}'$ that would result from additionally observing $\mathbf{x}$. In our setting, we take $\hat{p}$ to be uniform over the evaluation pool $\mathcal{X}_{\text{pool}}$, so the integral reduces to a sum over pool points.

{\noindent \em AF-III). Bayesian Active Learning by Disagreement (BALD):} 
BALD selects the points that maximize the mutual information between the predicted outcome and the model parameters 
\cite{houlsby2011bayesianactivelearningclassification}:
\begin{align}
    \begin{split}
            \alpha_{\text{BALD}}(\mathbf{x}) &= I(y;\, \omega \mid \mathbf{x}, \mathcal{D}_t) \\
    &= H[y \mid \mathbf{x}, \mathcal{D}_t] - \mathbb{E}_{p(\omega \mid \mathcal{D}_t)}\left[H[y \mid \mathbf{x}, \omega]\right]
    \end{split}
\end{align}
Intuitively, BALD seeks points where the surrogate model as a whole is uncertain in its prediction but individual posterior samples are confident on average---indicating genuine epistemic uncertainty rather than irreducible aleatoric noise.

{\noindent \em AF-IV). Expected Predictive Information Gain (EPIG):}
Whereas the BALD acquisition function seeks information gain about the surrogate model's parameters, EPIG calculates the expected mutual information between the label $y$ and a target prediction $y_*$ at a randomly sampled target input $\mathbf{x}_*$ \cite{smith2023predictionorientedbayesianactivelearning}.
\begin{align}
\begin{split}
    \alpha_{\text{EPIG}}(\mathbf{x}) &= \mathbb{E}_{p_{*}(\mathbf{x}_{*})}\left[ I(y;\, y_{*} \mid \mathbf{x}, \mathbf{x}_{*}, \mathcal{D}_t) \right] \\ 
    &= \mathbb{E}_{p_{*}(\mathbf{x}_{*})} \Bigl[ H[y_{*} \mid \mathbf{x}_{*}, \mathcal{D}_t] \\
    &\quad - \mathbb{E}_{p(y \mid \mathbf{x}, \mathcal{D}_t)}\left[ H[y_{*} \mid \mathbf{x}_{*}, \mathbf{x}, y, \mathcal{D}_t] \right] \Bigr]
\end{split}
\end{align}
In doing so, EPIG seeks to target information gain about the surrogate model's predictive performance directly.

For more information about the specific parameters of the surrogate models and acquisition functions that we used, please see the \hyperref[sec:Appendix]{Appendix}.

\subsection{Full Procedure}

The complete active testing procedure is summarized as follows. We initialize the surrogate model with $N_0$ evaluation points sampled uniformly at random from the design space to ensure broad initial coverage. At each subsequent round $t$, the GP surrogate is fit to all available evaluation data $\mathcal{D}_{t-1}$, the acquisition function is computed over all unevaluated candidate configurations, and the configuration(s) with the highest acquisition value are selected for real-world evaluation. After executing the policy under the selected configuration(s) and recording the performance outcome, the dataset is updated and the procedure repeats until the evaluation budget is exhausted. Please see Algorithm \ref{alg:active-testing}.

\begin{algorithm}[t!]
\begin{algorithmic}[1]
  \REQUIRE Target policy $\pi_\theta$, evaluation pool $\mathcal{X}_{\text{pool}}$, evaluation budget $T$, initial sample size $N_0$, acquisition function $\alpha$
  
  \STATE Sample $N_0$ configurations uniformly at random from $\mathcal{X}_{\text{pool}}$
  \STATE Execute $\pi_\theta$ under each configuration; collect initial dataset $\mathcal{D}_0 = \{(\mathbf{x}_i, y_i)\}_{i=1}^{N_0}$
  
  \FOR{$t = 1$ \textbf{to} $T - N_0$}
    \STATE Fit GP surrogate model with product kernel $k(\mathbf{x}, \mathbf{x}')$ on $\mathcal{D}_{t-1}$
    \STATE Compute acquisition function $\alpha(\mathbf{x})$ for all $\mathbf{x} \in \mathcal{X}_{\text{pool}} \setminus \mathcal{D}_{t-1}$
    
    \STATE Select next configuration $\mathbf{x}_t = \arg\max_{\mathbf{x}} \alpha(\mathbf{x})$
    
    \STATE Execute $\pi_\theta$ in the real world under configuration $\mathbf{x}_t$; obtain performance score $y_t$
    
    \STATE Update dataset: $\mathcal{D}_t = \mathcal{D}_{t-1} \cup \{(\mathbf{x}_t, y_t)\}$
  \ENDFOR
  
  \ENSURE Estimated performance profile $\{\mu_T(\mathbf{x})\}_{\mathbf{x} \in \mathcal{X}_{\text{pool}}}$ and dataset $\mathcal{D}_T$
\end{algorithmic}

\caption{Active Testing for Generalist Robot Policy Evaluation}
\label{alg:active-testing}
\end{algorithm}

\begin{figure}
    \centering
    \includegraphics[width=1\linewidth]{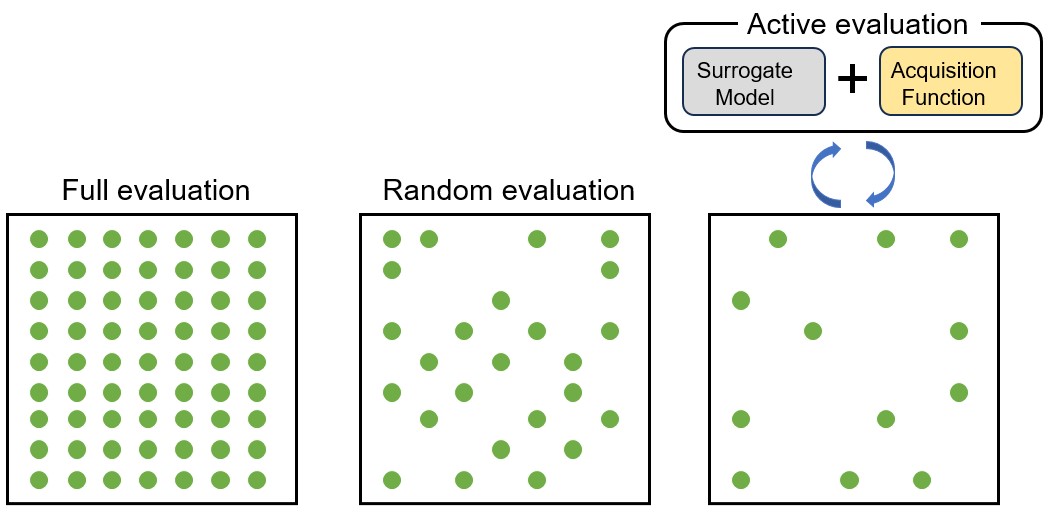}
    \caption[Comparison of evaluation strategies]{Comparison of evaluation strategies: full evaluation covers the entire space, random evaluation samples uniformly, and active evaluation selects informative points using a surrogate model and acquisition function.}
    \label{fig:eval_compare}
\end{figure}

\section{Experiments}
Limited by time or resources, existing works typically pick evaluation configurations by randomly sampling from the pool of 
possible configurations and report an aggregate, scalar success rate. This approach has several shortcomings. First, success 
rate is an oversimplification of policy performance; it does not capture the performance distribution across the space of 
possible configurations. This is especially important because the performance distribution indicates where demos should be 
collected next to finetune and improve the policy. Second, random evaluation can be accidentally biased towards high or 
low-performing areas; in the limited-evaluation regime, this is particularly harmful for accurate performance estimation. Our 
experiments compare the estimated performance distributions from active and random testing on a limited number of trials 
against the ground truth performance distribution.  A comparison
of full, random, and active evaluation can be found in Figure \ref{fig:eval_compare}. Note that, to the best of our knowledge, the current "state-of-the-art" in our setting is random testing and one specific active surrogate model and acquisition function combination \cite{anwar2025efficientevaluationmultitaskrobot}, Mixture Density Network (MDN) and Bayesian Active Learning by Disagreement (BALD), which we include in our experiments.

\begin{figure*}
    \centering
    \includegraphics[width=1.0\linewidth]{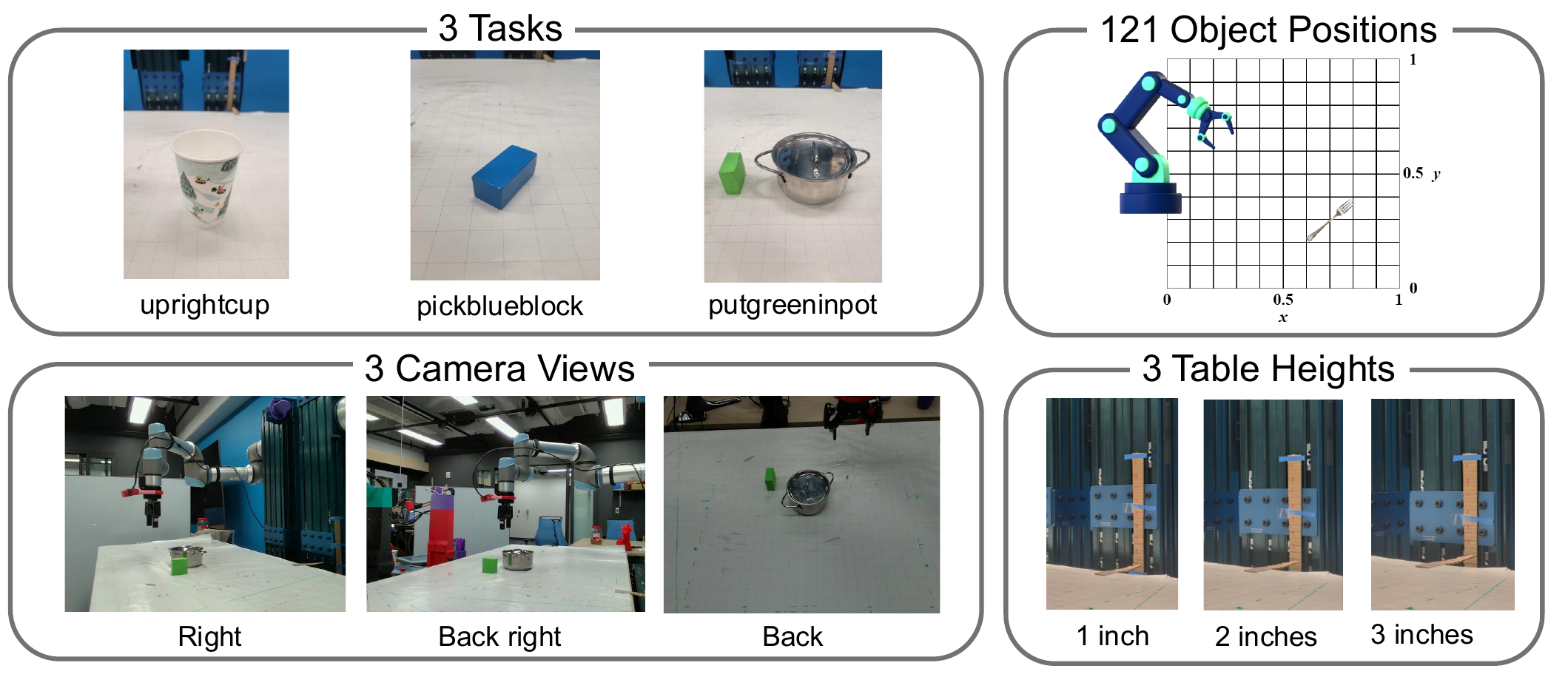}
    \caption[Evaluation factors]{Our evaluation suite: 3 tasks, 3 table heights, 3 scene camera positions, and 121 object positions on table.}
    \label{fig:setup}
\end{figure*}

\subsection{Experimental Details}
\subsubsection{Evaluation Methods}
The ground truth performance distribution is obtained via full evaluation, manually evaluating the target policy across the entire design space with real robot hardware while respecting reachability constraints of the UR5e arm. Our evaluation tasks and design space are illustrated in Figure \ref{fig:setup} and produce 2331 valid evaluation configurations: 804 total for the first two tasks and 723 for the third due to additional constraints from multiple objects (e.g., the initial positions of the block and lid, and the block and pot, should not overlap). Each of these evaluations took roughly 1 minute, which includes the robot execution and the operator initializing the scene to the next evaluation configuration. 

Active and random testing are done offline, sampling from these ground truth results to simulate the sequential testing process. In each trial, surrogate models are fit to both actively and randomly acquired evaluation data. Sampling was done without replacement and the stochasticity of the robot policy's outputs is assumed to be negligible based on empirical observations that the policy's behavior is consistent across repeated evaluations at fixed configurations. An illustration comparing full, random, and active evaluation can be found in Figure \ref{fig:eval_compare}.

For active testing, we "warm-start" the surrogate model with 30 initial randomly selected points. 

\subsubsection{Tasks}
We fine-tuned $\pi_0$, an off-the-shelf robot manipulation VLA model, to perform three real-world table-top manipulation tasks with a 
UR5e robot arm with visual observations coming from a scene and a wrist camera \cite{black2026pi0visionlanguageactionflowmodel}. More details on fine-tuning and demonstrations can be found in the \hyperref[sec:Appendix]{Appendix}.

The tasks are as follows:
\begin{description}
    \item \textbf{Pick up the blue block (pickblueblock):} A standard manipulation task testing the policy's ability to localize, reach, grasp, and lift a colored block from the table. 
    \item \textbf{Set the cup upright (uprightcup):} A more complex manipulation task requiring the policy model to implicitly understand the orientation of the object to recognize an "upright" goal state.
    \item \textbf{Put the green block in the pot (putgreeninpot):} A longer-horizon task that involves first taking off the lid on the pot.  The policy should visually discern between objects and execute subtasks in the appropriate order.
\end{description}

\subsubsection{Continuous Outcome Ranges}
Performance ($f$) for each task was graded on a discretized scale based on task progress (e.g., 0 to 6).
\begin{description}
    \item \textbf{Pick up the blue block (pickblueblock):}
    0.0: failed completely, 
    0.5: moved toward the block (within 10 cm), 
    1.0: moved to the block (within 5 cm), 
    1.5: tried to grab the block (touched it), 
    2.0: grasped block (success)
    \item \textbf{Set the cup upright (uprightcup):}
    0.0: failed completely,
    0.5: moved toward the cup (within 10 cm),
    1.0: moved to the cup (within 5 cm),
    1.5: tried to grab the cup (touched the rim),
    2.0: grabbed the cup,
    2.5: dropped the cup vertically ($\geq$45 degrees) onto its bottom rim,
    3.0: set the cup upright (success)
    \item \textbf{Put the green block in the pot (putgreeninpot):}
    0.0: failed completely,
    0.5: moved toward the lid (within 10 cm)
    1.0: moved to the lid (within 5 cm),
    1.5: tried to grab the lid (touched handle),
    2.0: grabbed lid and lifted it,
    2.5: dropped lid outside of the pot,
    3.0: moved toward the block (within 10 cm),
    3.5: moved to the block (within 5 cm),
    4.0: tried to grab the block (touched it),
    4.5: grabbed block,
    5.0: moved toward the pot (within 10 cm),
    5.5: moved block to the pot (within 5 cm),
    6.0: dropped the block in the pot (success)
\end{description}

\subsubsection{Factors}
The design space across all tasks consisted of three factors: 
\begin{description}
    \item \textbf{Object position (x, y):} The table was discretized into a $11 \times 11$ grid ranging from (0,0) to (1,1). Demonstrations for 
fine-tuning the robot policy were collected in the bottom left and top right quadrants from the perspective of the evaluator. 
Evaluation was done on the whole grid.
    \item \textbf{Table height:} Training demos were collected on a table height that was 1 or 3 inches from a reference height. 
Evaluation additionally included a table height of 2 inches.
    \item \textbf{Scene camera viewpoint:} The scene camera (as opposed to the wrist camera) was varied between three locations. Training demos were collected with the scene camera set either at the back or right of the scene 
or at the right. Evaluation used one additional scene camera viewpoint: back-right.
\end{description}

All factors not listed were kept constant (e.g., lighting, object orientations, initial position of the robot arm (small differences across training demos)).

\begin{figure*}
    \centering
    \includegraphics[width=0.96\linewidth]{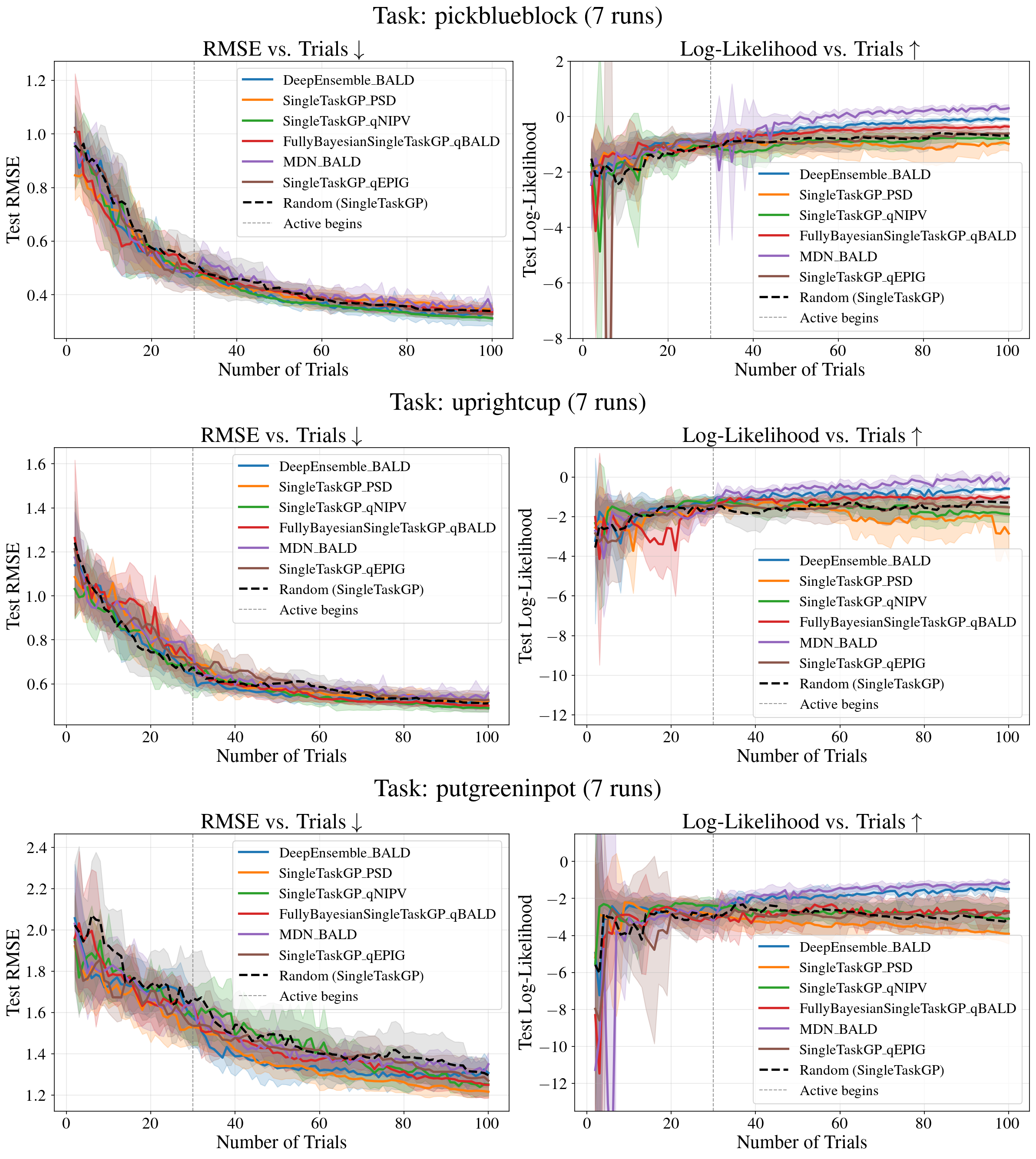}
    \caption[Active vs. random testing]{Active testing methods versus uniform-random testing across 3 tasks. Shaded regions indicate $\pm1$ standard deviation across runs. Active testing methods generally achieve better metrics than uniform-random in fewer trials.}
    \label{fig:active_vs_rand_alltasks}
\end{figure*}

\subsection{Efficiency of Active vs. Random Testing}
To quantify the efficiency of active testing relative to random testing, we measure how quickly each method approximates the ground truth 
performance distribution to a given level of accuracy.
We evaluate the performance of our proposed approach using both the root mean squared error (RMSE) and predictive log-likelihood metrics. 
Recall that our approach learns a probabilistic surrogate model mapping from task configuration to policy performance. RMSE evaluates the 
accuracy of the mean prediction of the surrogate, penalizing outliers to indicate performance across the design space. In contrast, 
predictive log-likelihood evaluates the quality of the predictive probability distribution, rewarding both good mean predictions and 
well-calibrated uncertainty of the model. 

Figure \ref{fig:active_vs_rand_alltasks} shows RMSE and log-likelihood on the evaluation dataset as a function of policy evaluations (trials). We make the following conclusions:
\begin{itemize}[leftmargin=*]
\item Across all three tasks, uniform random testing is a strong baseline but most active testing variants match or exceed it on both RMSE and log-likelihood. In terms of RMSE, the surrogate with active testing typically matches the best value achieved by the random-testing surrogate in 20-40 fewer trials (out of 100).\footnote{For random testing specifically, we use a Gaussian Process surrogate but because the method does not rely on uncertainty estimates, one could use any regression model.} In terms of log-likelihood, the active surrogate typically matches the best value achieved by the random surrogate in 50-65 fewer trials. This suggests that active testing improves the mean prediction and also provides better-calibrated uncertainty for the surrogate. 
\item For both metrics, we observe lower variances across runs for active testing methods compared to random testing. In other words, active methods sample similar points across runs due to both the surrogate model and acquisition function. Lower variance allows the evaluator to be more confident that in any given run, their surrogate model will converge consistently. This is an important consideration because evaluation is limited and usually run just once.
\item The choice of surrogate model and acquisition function yielded different relative
performances in RMSE and log-likelihood. To determine the best-performing combination, we changed surrogate models while keeping acquisition function the same and vice versa. These isolated comparisons are given in the \hyperref[sec:Appendix]{Appendix}. Both metrics considered, Deep Ensemble + BALD is the best choice.
\item RMSE converges to a nonzero value for all methods. Apart from an imperfect model and limited training data, this might be indicative of aleatoric noise, irreducible noise from the data-generating process (e.g., inconsistencies with labeling, stochasticity of the policy). This is supported by ablations with and without a noise floor constraint in the surrogate; having a small noise floor helped the surrogate converge faster.
\end{itemize}

\subsection{Generalization}
\begin{figure}
    \centering
    \includegraphics[width=1.0\linewidth]{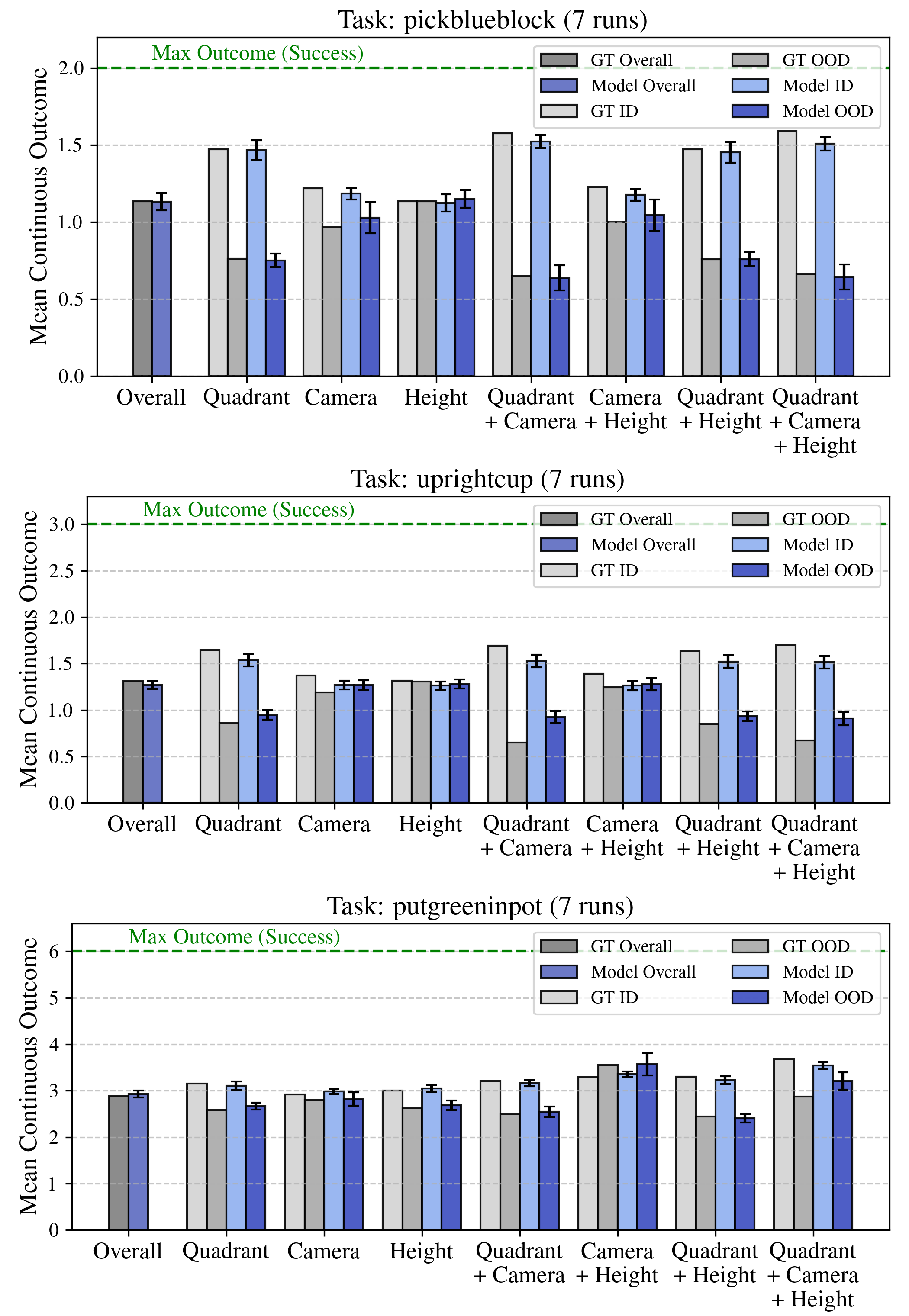}
    \caption[Generalization of $\pi_0$]{Ground truth and surrogate model-predicted mean outcome for ID \& OOD factors. The surrogate model used here is the final model from active or random evaluation (trial 100). It can accurately model generalization ability with low variance.}
    \label{fig:id_ood_alltasks}
\end{figure}

A primary concern in model evaluation is characterizing the generalization of the model to out-of-distribution (OOD) 
conditions. Evaluating ID and OOD performance also speaks to the fairness of testing methods, i.e. how well the sampled test points allow 
us to model \textit{all parts} of the performance distribution. Having good, consistent predictive performance of the surrogate model 
everywhere in the design space is crucial in the limited-evaluations regime where randomness can lead to biases in sampling.
In Figure \ref{fig:id_ood_alltasks} we show the ground truth and surrogate-predicted performance of the robot policy on in- and out-of-distribution factor values (top left and bottom right quadrants, table height 2, back-right scene camera viewpoint). We make the following conclusions:
\begin{itemize}[leftmargin=*]
    \item A surrogate model trained on evaluation results can accurately predict both ID and OOD performance for single and multiple factors, offering a sample-efficient way to fairly evaluate generalization of the policy.
    \item Among our single-factor variations across tasks, policy performance was consistently the most sensitive to object position (x, y), slightly less for camera viewpoint, and least for table height. This is in line with a recent work \cite{saxena2025matterslearninglargescaledatasets, gao2024efficientdatacollectionrobotic}. That being said, it is important to note the amount that each factor was changed since one could conclude that policy performance is actually not sensitive to object position if they only change the position by a small amount. The distance from the in-distribution camera viewpoints (back, right) to the out-of-distribution camera viewpoint (back-right) was about 2.5 feet. The distance from the ID table heights (1, 3) to the OOD table height (2) was 1 inch. The distance from an ID quadrant (bottom left, top right) to an OOD quadrant (top left, bottom right) was about 6 inches.
    \item Multi-factor variations have an additive impact on outcome (e.g., $\Delta$ performance for OOD camera viewpoint and table height is roughly equal to the $\Delta$ performance for OOD camera viewpoint + $\Delta$ performance for OOD table height). 
    \item Some regions of the factor space are "harder", requiring more demonstrations than others to perform well. For example, we observe that the policy performs much worse for table height 1 than for table height 3 although both have an equal number of demonstrations.
\end{itemize}

\section{Conclusion}
In this work, we present a sample-efficient, factor-based active evaluation framework for generalist robot manipulation policies that treats 
policy evaluation as an active experimental design problem. For evaluation, we use the Bayesian active testing framework to define a probabilistic surrogate model over the design space of task factors and use principled acquisition functions to guide evaluation selection. Our approach enables accurate characterization of policy performance in at least 20-40\% fewer evaluation trials than random testing.

\subsection{Limitations \& Future Work}
There are multiple avenues for future work in this problem space. The benefits of active testing compared to random testing might be more pronounced with more complex performance distributions (e.g., with more factors), wider factor ranges, and wider outcome ranges (our task outcomes at most ranged from 0 to 6). Additionally, more expressive representations or encodings for the design space and surrogate model architectures could yield stronger surrogate models. For instance, one could use the existing pretrained vision-text encoder from the robot model and add a prediction head. Lastly, one could consider transfer learning opportunities (e.g., adapting a surrogate model trained on some tasks to a new task).

In the data curation experiment (see \hyperref[sec:Appendix]{Appendix}), we had a couple of curious observations that would benefit from further analysis. For instance, we found that the points with worse outcomes do not make for more optimal points to collect extra demonstrations for. Furthermore, we found that some regions in the factor space are harder than others to increase outcomes for and some points actually experience a drop in outcome despite the robot policy having extra training demonstrations. In addition to gaining more insights about the relationship between factors and generalization, explaining these observations would help distinguish between hard regions and uninfluential regions at which to collect demonstrations; using only outcomes, these two are currently indistinguishable.

Furthermore, using more sophisticated methods like influence functions \cite{agia2025cupidcuratingdatarobot} or a combined active-testing and active-learning approach \cite{yu2023actively} might allow for more optimal selection strategies. However, these works will require some adaptation as their inputs and outputs do not match our setting. The influence function work uses state action pairs from collected demos as input instead of factor values of potential demos. The combined active-testing and active-learning work assumes the output of the main model (the policy in our case) is the same as the output of the surrogate model.

\bibliography{refs_arxiv}

\clearpage
\section{Appendix} \label{sec:Appendix}

\subsection{Policy Finetuning}
\begin{itemize}[leftmargin=*]
\item The following numbers of demonstrations were collected manually to finetune $\pi_0$ for each task: 120 for "uprightcup", 136 for "pickblueblock", and 160 for "putgreeninpot.
\item Each $\pi_0$ model was finetuned with LoRA for 30000 steps following the steps in the official "openpi" repository. 
\end{itemize}

\subsection{Active Testing Components}
\subsubsection{Best Combinations}
 The full comparison of all combinations of surrogate models and acquisition functions can be found in Figure \ref{fig:active_vs_rand_alltasks}. Here, we show ablations of either only acquisition functions or only surrogate models to highlight the best combination for sample-efficient evaluation.

\begin{figure*}
    \centering
    \includegraphics[width=1\linewidth]{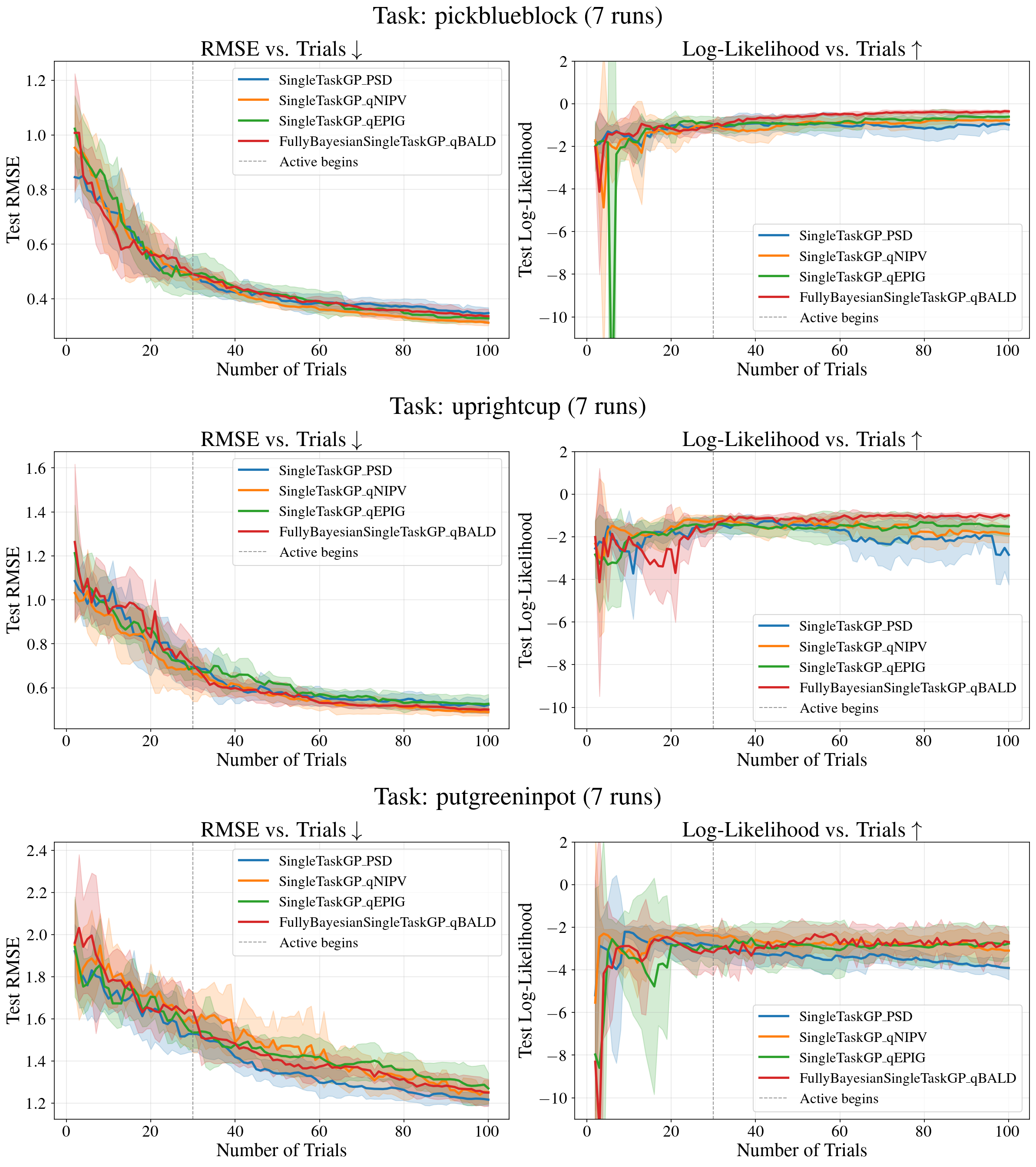}
    \caption[Acquisition function comparison]{Comparison of different acquisition functions using the same surrogate model (Gaussian Process). BALD is superior in log-likelihood and competitive in RMSE. As before, shaded regions indicate $\pm1$ standard deviation across runs.}
    \label{fig:acqf_ablation}
\end{figure*}

Figure \ref{fig:acqf_ablation} shows an ablation of acquisition functions, keeping the surrogate model (Gaussian Process) the same. Bayesian Active Learning by Disagreement (BALD) is superior in log-likelihood and competitive in RMSE whereas variance-maximizing functions like Posterior Standard Deviation (PSD) and Negative Integrated Posterior Variance (NIPV) are competitive on RMSE but not so on log-likelihood.

\begin{figure*}
    \centering
    \includegraphics[width=1\linewidth]{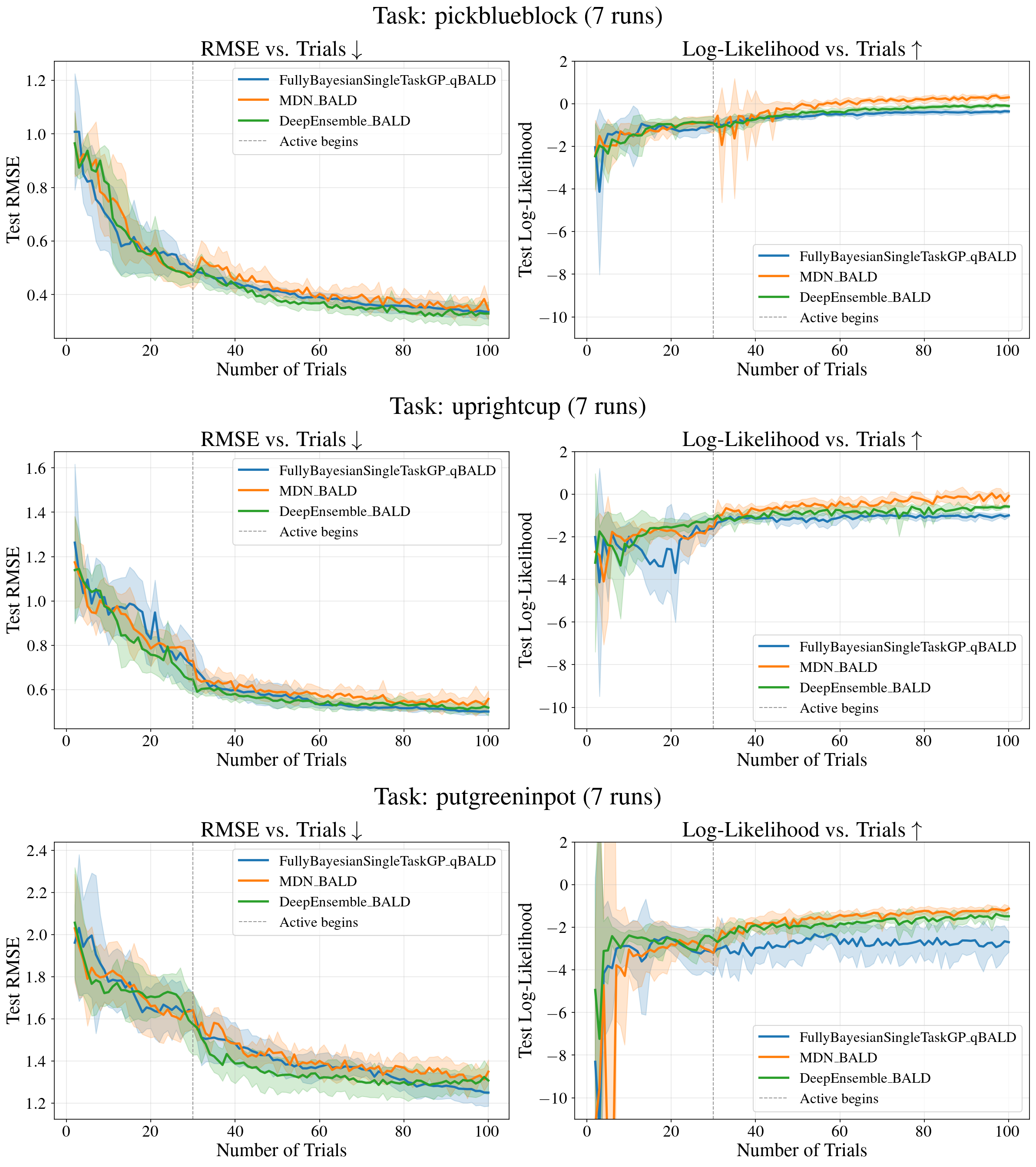}
    \caption[Surrogate comparison]{Comparison of different surrogate models using the same acquisition function (Bayesian Active Learning by Disagreement). Deep Ensemble converges the fastest in terms of RMSE while being competitive in log-likelihood. As before, shaded regions indicate $\pm1$ standard deviation across runs.}
    \label{fig:surrogate_ablation}
\end{figure*}

Figure \ref{fig:surrogate_ablation} shows an ablation of surrogate models, keeping the acquisition function (BALD) the same. Deep Ensemble converges the fastest in terms of RMSE while being competitive in log-likelihood whereas Gaussian Process (GP) and Mixture Density Network (MDN) are competitive in only RMSE or log-likelihood respectively.

Thus, we conclude that Deep Ensemble + BALD is the best-performing combination.

\subsubsection{Surrogate Model}
\begin{figure*}
    \centering
    \includegraphics[width=1\linewidth]{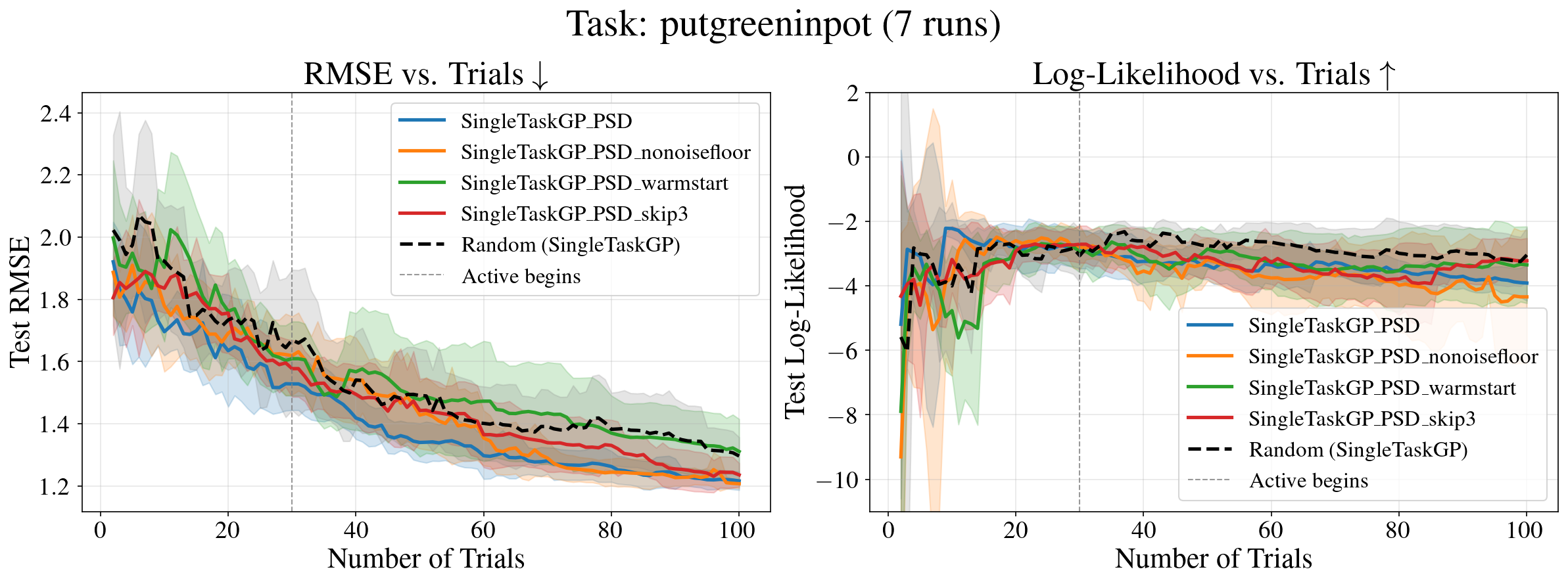}
    \caption[Gaussian Process design choice ablation]{Several Gaussian Process design choices: 1) standard (noise floor of $10^{-1}$), 2) no noise floor, 3) warm start, 4) refitting every three trials instead of every trial. The configuration using a noise floor converges the fastest.}
    \label{fig:putgreeninpot_ablation}
\end{figure*}

\begin{itemize}[leftmargin=*]
  \item \textbf{Gaussian Process:} For the Gaussian Process ("SingleTaskGP" on BoTorch), we use a noise floor of $10^{-1}$ because of possible aleatoric noise (e.g., human labeling inconsistency, policy stochasticity). Figure \ref{fig:putgreeninpot_ablation} shows the ablation of several model design choices: 1) no noise floor, 2) warm start (when refitting at trial $n$, start from the previous trial's model parameters), and 3) refitting every three trials instead of every trial. Using a noise floor of $10^{-1}$ converges the fastest.
  \item \textbf{Fully Bayesian Gaussian Process:} The Fully Bayesian Gaussian Process surrogate ("FullyBayesianSingleTaskGP" on BoTorch) uses Markov Chain Monte Carlo (MCMC) with 128 warm-up steps and 128 samples.
  \item \textbf{Mixture Density Network:} For the MDN, we use 2 Gaussian distribution components and the MLP has 4 hidden layers of dimension 32 and dropout probability of 0.1. The model was trained for 200 epochs.
  \item \textbf{Deep Ensemble:} For the Deep Ensemble, we use an ensemble of 4 MLPs each with 4 hidden layers of dimension 32 and dropout probability of 0.1. The model was trained for 200 epochs.
\end{itemize}

\subsubsection{Acquisition Function}
\begin{itemize}[leftmargin=*]
    \item \textbf{Expected Weighted Disagreement (XWED):}
XWED, introduced as an alternative to BALD for the active testing problem \cite{kossen2022activesurrogateestimatorsactive}, 
weights each BALD disagreement term by the loss of the original model $f$ at the inputs $x$ compared to the ground truth $y$:
\begin{align*}
    \alpha_{\text{XWED}}(x) &= \mathbb{E}_{Y \sim \pi(\cdot|x)}[-\mathcal{L}(Y, f(x)) \log \pi(Y|x)] \\
    &\quad - \mathbb{E}_{\Theta \sim \pi(\cdot)} \Bigl[ \\ 
    &\qquad \mathbb{E}_{Y \sim \pi(\cdot|x, \Theta)} [-\mathcal{L}(Y, f(x)) \log \pi(Y|x, \Theta)] \Bigr]
\end{align*}
The intuition is that weighting by the loss targets points that would contribute significantly to the final risk estimate.
However, XWED is inappropriate for our problem for two reasons:
\begin{itemize}[leftmargin=*]
    \item The output of the generalist model (actions) differs from the output of the surrogate model (outcomes). Therefore the loss term 
is not defined.
    \item Instead of building an accurate \textit{mean} risk estimate, we care more about accurately modeling the outcome at every point. XWED prioritizes 
the points with the most loss, leading to highly imbalanced coverage across the points.
\end{itemize}
\end{itemize}

\subsection{Evaluation}
\begin{itemize}[leftmargin=*]
    \item 121 possible object positions were reduced to 86-93 positions depending on the reachability of the UR5e arm at different table heights.
    \item To avoid unnecessarily prolonged evaluations, a maximum number of steps was set to complete each task, provided the UR5e arm did 
not trigger an E-stop (e.g., colliding with an object): 15 ($\approx$30 seconds) for "pickblueblock", 25 for "uprightcup" 
($\approx$50 secs), 35 for "putgreeninpot" ($\approx$1 min). These numbers also roughly reflect the relative complexity of each task; 
for example, successful trajectories typically took ($\approx$7 steps) for the "pickblueblock" task but ($\approx$17 steps) for "putgreeninpot".
\end{itemize}

\subsection{Targeted Data Collection}
Recent advances in data-driven robot learning have been largely driven by large amounts of training data in the form of human-collected demonstrations \cite{embodimentcollaboration2025openxembodimentroboticlearning}. However, since data collection for robotics is time-intensive, especially on real hardware, recent works study how to choose the demonstrations that will improve downstream policy performance the most \cite{chen2025curatingdemonstrationsusingonline,shi2025diversityneedscalablerobotic}. Some methods explicitly leverage compositional generalization to reason about how much coverage is needed along different factors \cite{gao2024efficientdatacollectionrobotic, zha2025guidingdatacollectionfactored}. Another set of methods rely on heuristics or proxy objectives to prioritize high quality demonstrations \cite{hejna2025robotdatacurationmutual} and still others try to estimate the downstream performance influence of training data \cite{agia2025cupidcuratingdatarobot,dass2025datamilselectingdatarobot}. In this section, we explore how to use the surrogate model trained from evaluation data to select factor configurations to collect demos for which are used to further finetune a robot policy. We compare against a naive baseline and \footnote{We use "data curation" and "targeted data collection" interchangeably.}

The baseline and novel strategies are as follows:
\begin{enumerate}
    \item \textbf{Observed Failures:} Simply collect new demos at all the observed failure points or factor configurations.
    \item \textbf{Certain Failures:} We use the final surrogate model from active testing to choose predicted failure points, or factor configurations, (low outcomes with high certainty) across the \textit{entire design space}. Specifically, we use the surrogate model to select the points with the lowest predicted outcome in the 30th percentile of predicted variance. Therefore, we call this method "Certain Failures". Importantly, this method can potentially select points with a worse mean outcome than Observed Failures can since these points need not have been visited during testing.
\end{enumerate}
    
For each strategy, we collect new demonstrations, add them to the original dataset, fine-tune the model on this augmented dataset, and conduct full evaluation. We did this experiment for the "uprightcup" and "putgreeninpot" tasks.

The experiments for each of the two tasks differed slightly in their factor design space to test the methods at two levels of data complexity (more factor variation, more complex performance distributions). With a larger factor space, active methods for evaluation and data collection might show a larger advantage than with a smaller space.

The "uprightcup" experiment used a small factor design space, fixing ($x$, $y$) positions to the top left quadrant and table height to 1, giving a total of 108 configurations. The "putgreeninpot" experiment used a larger factor space, fixing ($x$, $y$) positions to the bottom left and top left 
quadrants, table height to 1 and 3, and camera viewpoint to the "back" and "right" positions, giving a total of 218 configurations.

For fairness of comparison, all methods tested in each experiment were based on the same active testing run and used to select the same number of points. For instance, in the "putgreeninpot" experiment we observed 24 failure points in the active testing run we selected at random (the maximum possible points for the "Observed Failures" baseline method) so "Certain Failures" was used to select the same number of points.

\begin{figure}
    \centering
    \includegraphics[width=1\linewidth]{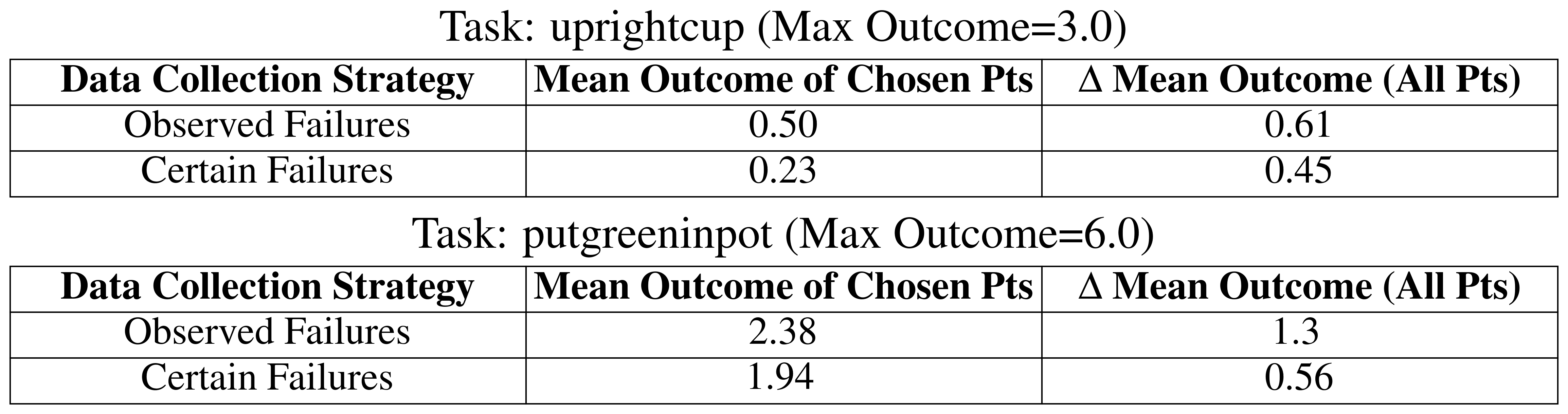}
    \caption{Mean continuous outcomes after using either a surrogate prediction-based data collection strategy versus collecting only at failure points observed during evaluation. "Mean Outcome of Chosen Pts" refers to the outcomes of just the points obtained via full evaluation prior to collecting demos for them and adding them to the training data. "Change in Mean Outcome (All Pts)" refers to the outcomes over the entire factor space after training on the augmented dataset. Despite selecting points with higher mean outcome than the other methods, Observed Failures drives the largest positive change in mean outcome compared to the other methods.}
    \label{fig:curation_comparison}
\end{figure}

Figure \ref{fig:curation_comparison} shows the baseline mean outcome of the points that are selected for demo collection, and the resultant mean outcomes across the entire factor space after fine-tuning and running full evaluation. We make the following observations: 
\begin{itemize}[leftmargin=*]
    \item Across both experiments, both data collection methods improved mean outcome but Observed Failures improved mean outcome more. 
    \item Observed Failures boosts mean outcome the most despite choosing points with higher outcomes on average to collect demonstrations for. This suggests that purely collecting demonstrations at the worst possible points is not optimal and there are generalization effects that are unaccounted for.
\end{itemize}

\begin{figure}
    \centering
    \includegraphics[width=1\linewidth]{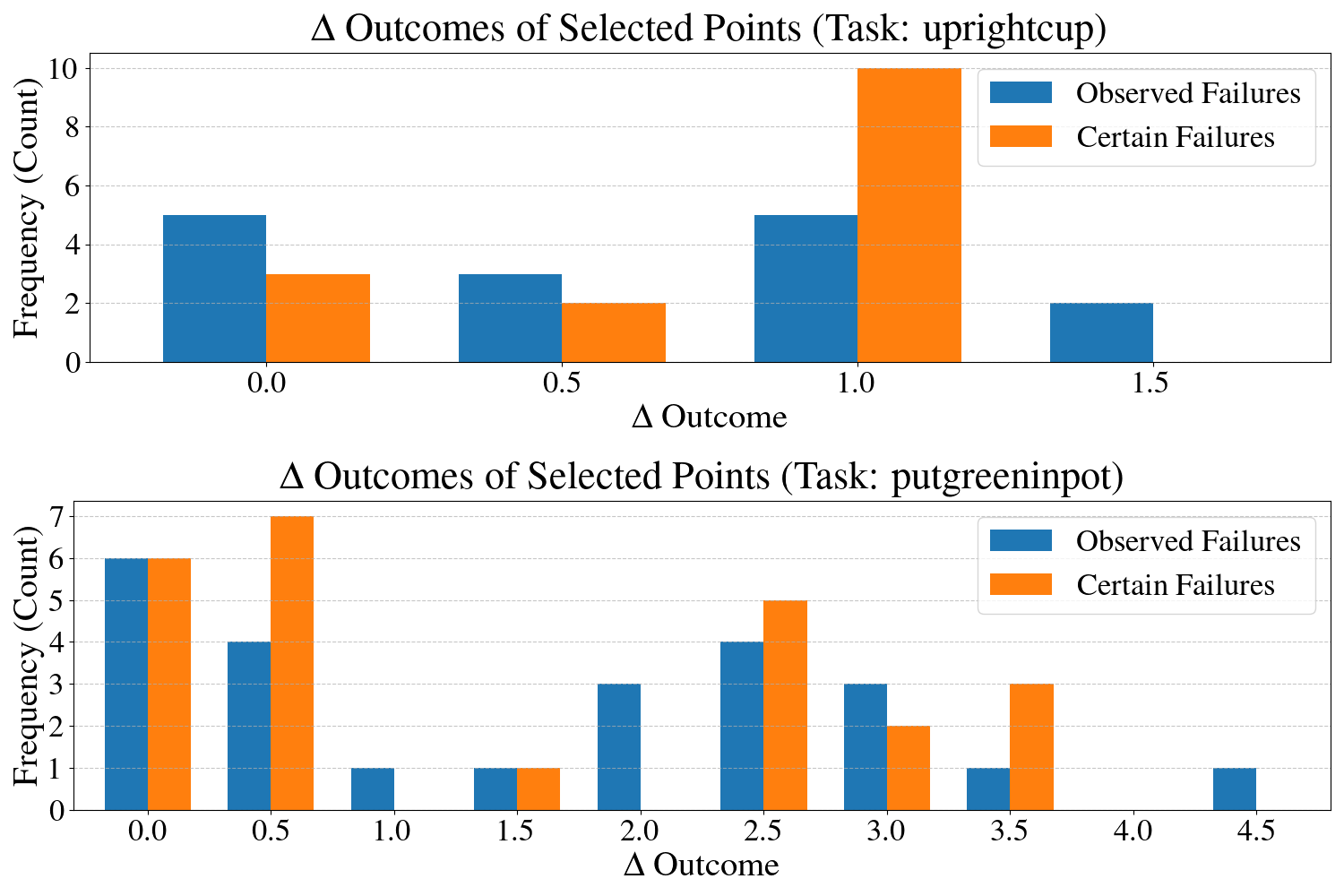}
    \caption[Change in outcome of curated points]{Change in outcomes of just the points selected for extra demo collection for each data curation experiment. Nearly all points have a non-negative change in outcomes, confirming that demonstrations improve outcomes where they are collected.}
    \label{fig:points_delta_outcomes}
\end{figure}

Figure \ref{fig:points_delta_outcomes} shows the changes in outcome for the points selected by the data curation methods. Nearly all points have a non-negative change in outcomes, confirming that demonstrations improve outcomes where they are collected. Additionally, because the change in outcome at the selected points was very similar between methods, the advantage of Observed Failures in boosting mean outcome compared to other methods is primarily driven by the boost in outcomes for points in the factor space that were \textit{not} selected. Thus, the points that Observed Failures chooses are more "influential" than Certain Failures i.e. demonstrations collected at these points generalize better to the other points in the factor space.

\begin{figure}
    \centering
    \includegraphics[width=1\linewidth]{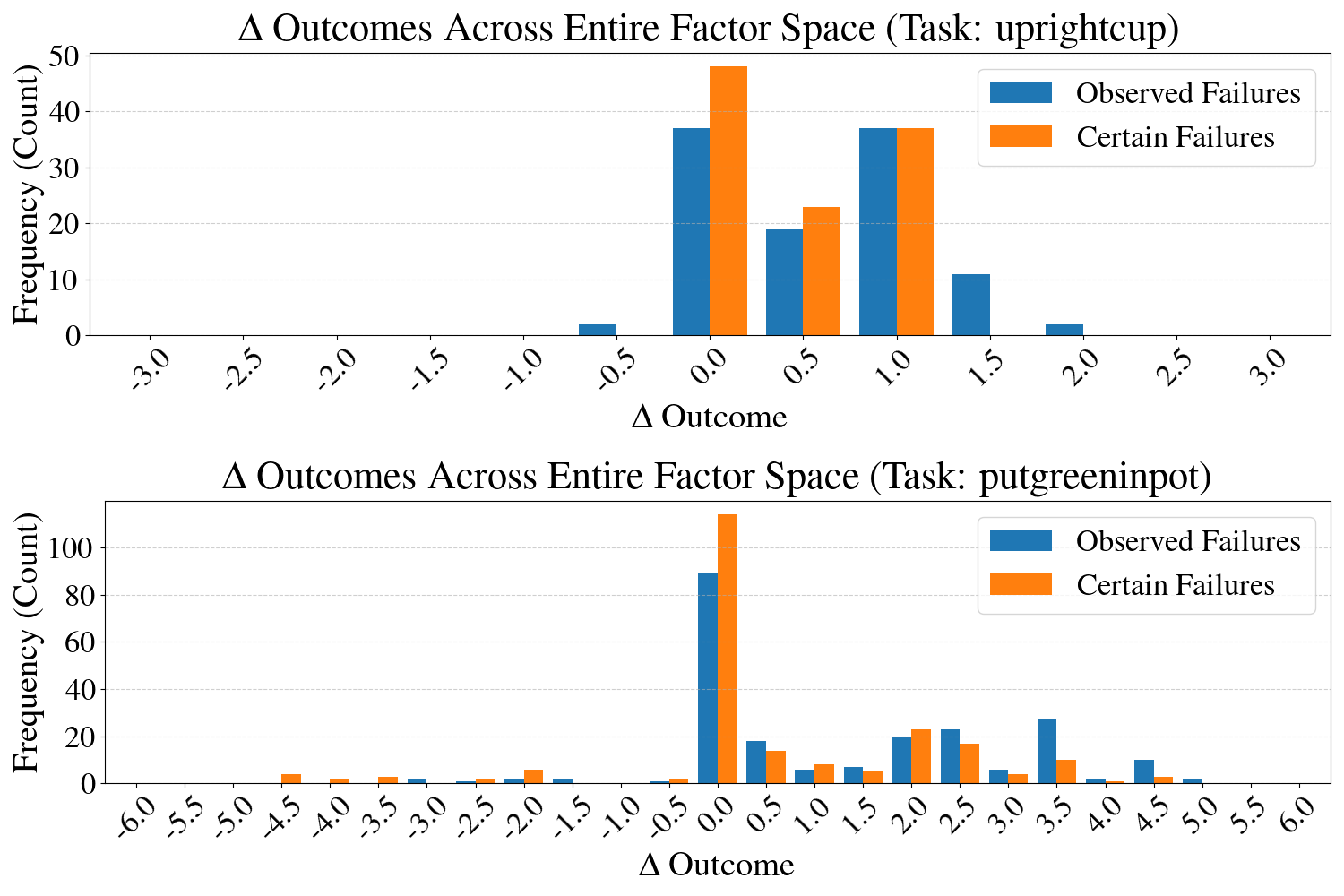}
    \caption[Change in outcome across factor space]{Change in outcomes of all points in the factor spaces for each data curation experiment. Most points have either no change or an increase in outcomes (typically expected with more training data) while a few have a decrease, showing that generalization exists only weakly for the considered factors.}
    \label{fig:delta_outcomes_full_factor_space}
\end{figure}

Figure \ref{fig:delta_outcomes_full_factor_space} shows the changes in outcome for all points in the factor spaces after collecting the extra demonstrations and conducting full evaluation. Most points do not experience an increase in outcome, showing that generalization exists narrowly for the considered factors. We find that although all methods' chosen points had positive changes in their outcomes after fine-tuning, there are a few other points in the factor spaces that had negative changes in their outcomes. Since the number of training demonstrations increases in these data curation experiments, this suggests that a change in the distribution of factors where the training demos are collected can hurt policy performance in some regions of the factor space. In other words, collecting more in one region of the factor space may hurt performance in another.

\begin{figure}
    \centering
    \includegraphics[width=1\linewidth]{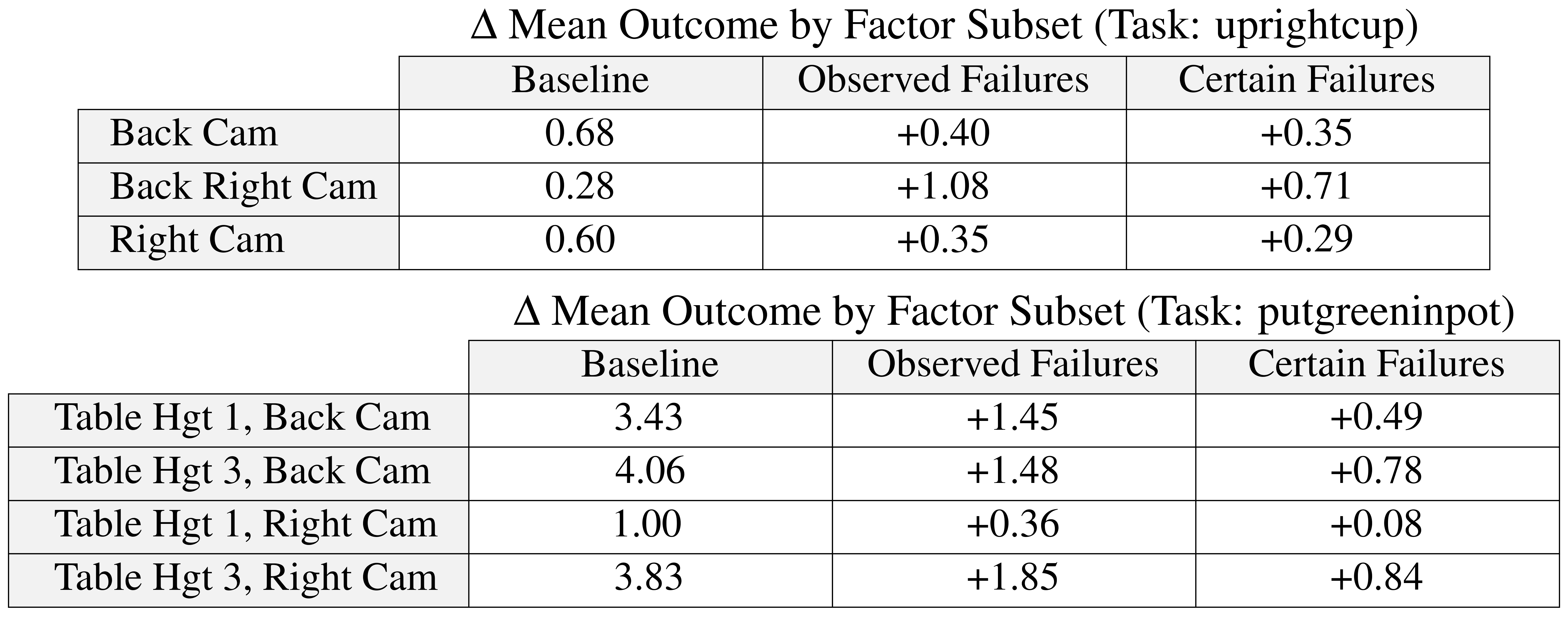}
    \caption[Data curation impact by factor subset]{Baseline outcome (before data collection), number of points collected, and change in mean outcome by factor subset for each data curation experiment (OF=Observed Failures, CF=Certain Failures). All subsets' mean outcomes increased across both methods but some subsets experienced an outsized increase relative to the number of points the method selected in that region.}
    \label{fig:delta_outcomes_by_subset}
\end{figure}

We also reinforce our finding from the previous experiment that some regions of the factor space require more demonstrations than others to perform well. Figure \ref{fig:delta_outcomes_by_subset} shows the original mean outcomes and the changes in mean outcomes across the factor space for the unique factor subsets in each data curation experiment. For instance, in the "uprightcup" experiment, the table height (1) and quadrant (top left) stay fixed so the variable subsets are the camera viewpoints. In the "putgreeninpot" experiment Certain Failures selects most of its points from table height 1, yet from the figure, we see that the mean outcome across the factor space increases much less at table height 1 than it does for table height 3.

\end{document}